# Bilateral Network with Residual U-blocks and Dual-Guided Attention for Real-time Semantic Segmentation

1st Liang Liao  2nd Liang Wan*  3rd Mingsheng Liu
*State Key Laboratory of Public Big Data*
*College of Computer Science and Technology ,Guizhou University*
Guiyang, China
1st liaoliang.liang@163.com
2nd * lwan@gzu.edu.cn

4th Shusheng Li
*School Of Computer Science And Engineering, Northeastern University*
Shenyang, China
4th 765926064@qq.com

*Abstract*—When some application scenarios need to use semantic segmentation technology, like automatic driving, the primary concern comes to real-time performance rather than extremely high segmentation accuracy. To achieve a good trade-off between speed and accuracy, two-branch architecture has been proposed in recent years. It treats spatial information and semantics information separately which allows the model to be composed of two networks both not heavy. However, the process of fusing features with two different scales becomes a performance bottleneck for many nowaday two-branch models. In this research, we design a new fusion mechanism for two-branch architecture which is guided by attention computation. To be precise, we use the Dual-Guided Attention (DGA) module we proposed to replace some multi-scale transformations with the calculation of attention which means we only use several attention layers of near linear complexity to achieve performance comparable to frequently-used multi-layer fusion. To ensure that our module can be effective, we use Residual U-blocks (RSU) to build one of the two branches in our networks which aims to obtain better multi-scale features. Extensive experiments on Cityscapes and CamVid dataset show the effectiveness of our method. On Cityscapes, our light version network without pretrain weight can achieve 71.1% mIoU at 163 FPS on a single Nvidia RTX 3070 using full resolution images(1024×2048pix). And the large version can achieve 77.9% mIoU with a speed of 43 FPS which still reaches the real-time criterion. Our code and module has been open sourced at https://github.com/LikeLidoA/BiDGANet/tree/main

## I. INTRODUCTION

Sematic Segmentation is a fundamental task in computer vision, which aims to divide images into semantically meaningful regions. It has been widely applied in a variety of fields, such as autonomous driving, computational photography and human-machine interaction. Since the fully convolutional network(FCN)[1] was proposed to solve the problem of image segmentation, deep learning technologies have started to surpass traditional methods based on handcrafted features in terms of accuracy and efficiency. At the same time, some application scenarios have highly real-time requirement of inference speed. The methods like Deeplab[2], Segnet[3], PSPnet[4] can achieve encouraging segmentation accuracy but consume excessive memory space and sacrifice time complexity when it comes to high-resolution inputs. It significantly restricts their application to real-time cases like autonomous driving, which typically needs high resolution of input images and very low output latency.

In recent years, outstanding improvement has been achieved by a lot of works[5][6][7] in the field of real-time semantic segmentation, especially the autonomous driving scene. These works make efforts on the trad-off between segmentation accuracy and inference speed and the goal is to improve the usability of the model on devices without high computing capacity. These methods fall roughly into two categories. One is the single branch encoder-decoder architecture represented by [8][9], they directly follow the research line since FCN was proposed. Another category is the two-[6] or multi-branch[5] architecture which are specifically designed for real-time semantic segmentation. The difference between the two categories is the way they deal with multi-scale semantic features. Most encoder-decoder methods use layer-by-layer down-sampling and feature fusion operations to extract semantic features, which means the entire process starts and finishes in a single branch. By contrast, the multi-branch methods rethink the low-level details and high-level semantics, they suggest that spatial details and context semantics can be extracted separately. Especially, since the totally hand-craft two-branch network named BiSeNetV2[7] achieved state-of-the-art performance, two-branch architecture has become a representative paradigm for real-time semantic segmentation. The two-branch architecture not only achieves better boundaries and small objects segmentation performance than the single branch encoder-decoder architecture, but also realizes faster inference speed.

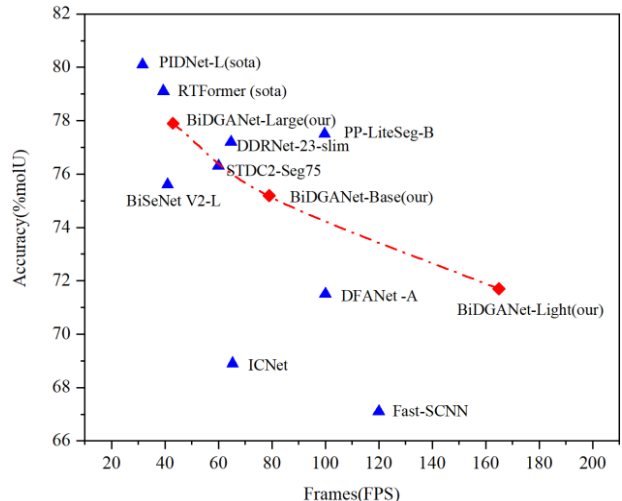

Fig. 1. **A comparison of speed-accuracy trade-off on Cityscapes.** The red rhombuses indicate our methods while blue triangles represent other methods.

In order to accelerate down-sampling process and reduce the memory access cost, features are extracted separately at different resolutions till the end, in most two-branch networks it leads to a heavy feature fusion at the last part of networks. Besides, famous networks such as BiSeNetV2 cannot avoid using hand-craft lightweight backbone, this also limits its performance.

In this paper, we propose a new two-branch network which inherits the classic design concept and makes improvements in feature fusion. To improve the ability of extracting multi-scale features in our network, we introduce the excellent network called U$^2$-Net[11] from the field of Salient Object Detection (SOD). We mainly use the basic construct of this network called Residual U-blocks (RSU) to build the low-resolution branch which actually is the backbone of our network. The other branch in our network we named high-resolution branch inherits the detail branch in BiSeNetV2. Based on the two branches, we propose the Dual-Guided Attention (DGA) module to fuse the feature during the last part of encoding. All the works above makes our network present the structure of heavy encoding and light decoding. This is the key to our network which has the ability to pursue real-time performance. Finally, our network achieves quite competitive results on two standard benchmarks: Cityscapes[12] and CamVid[13].

We summarize the contributions of this paper to the following points:
- We introduce a powerful module from U$^2$-Net to build the low-resolution branch of our network, which greatly improve the multi-scale feature extraction capability of our network.
- We propose a lightweight attention mechanism named Dual-Guided Attention (DGA) to realize feature fusion and attention computation simultaneously.
- Based on all efforts above, we build a real-time two-branch segmentation network named BiDGANet and achieve competitive results on the standard benchmarks.

## II. RELATED WORK

Real-time semantic segmentation networks have attracted much attention because of the growing demand for practical applications. To achieve the trade-off between inference speed and accuracy, many researches have contributed their efforts.

### A. Single-branch encoder-decoder architecture

Some methods choose classic Encoder-Decoder as major structure. They use layer-by-layer down-sampling and feature fusion operations to extract semantic features, which means they encode the low-level details and high-level semantics simultaneously. ESPNet[8] proposed to use parallel convolutions with different dilation rates to increase the receptive field in order to increase the efficiency of decoder. EDANet[9] proposed EDA module that input images and output features are densely connected within a larger block so that information can be shared across a wider receptive field. DFANet[10] employed deeply multiscale feature aggregation and lightweight depthwise separable convolutions that effectively refine high-level and low-level features. Although the methods above used to achieve state-of-the-art performance, most methods run at a slow inference speed when it comes to higher resolution input images.

### B. Two branch architecture

Different from the single-branch architecture methods, two-branch architecture has been proposed to preserve high-resolution details extracted early in the network by independently extracting features at different scales. BiSeNet[6] proposed a two-branched architecture consisting of a context path and a spatial path. The former based on a compact pre-trained backbone aims to extract contextual information while the latter utilizes a few convolutional layers to focus on the spatial details. After that, BiSeNetV2[7] further simplified the network structure. It proposed Bilateral Guided Aggregation to replace the Feature Fusion Module in BiSeNet and designed a totally hand-craft Semantic Branch. All efforts above made the network more efficient. Nevertheless, lacking effective feature interaction between the branches partly causes the accuracy decline, all the features are fused at the end of network, resulting in a new bottleneck. Hence, Fast-SCNN[14] and DDRNet[15] adopted a shared-backbone architecture. They start their networks from one branch and then divide into two parallel deep branches with different resolutions. Such design shares a part of network parameters early in the network, allowing them to add many interactions in the middle of the network, such as Bilateral Fusion proposed by DDRNet. But, they just avoide rather than solve the problem that to fuse the features of two branches actually becomes a performance bottleneck for many two-branch models.

### C. Attention mechanism

Attention mechanisms are used to solve local problems of neural networks, it can associate local information with the global to select the information that requires more attention. DANet[20] proposed Dual Attention, they employed position attention module to focus spatial information and channel attention module to correlate channel information. However, due to the channel attention module has high computational cost, if the backbone output feature maps with deep channels, it will not be suitable for real-time circumstance. Hence, CCNet[21] only considered spatial information and proposed criss-cross attention module to find connections between pixels in other positions in the same row or column. It reduces the amount of computation, making it possible to stack multiple attention modules. Besides, linear complexity attention mechanism such as External Attention[17] can achieve results comparable to the self-attention mechanism[22] with a small computational cost. These efforts became the motivation for our design, prompting us to design Dual-Guided Attention for our two-branch architecture network.

## III. METHOD

In this section, we elaborate the details of our work. Fig. 2 explain the construction of our network. Detailed descriptions of each component are below.

### A. High-resolution Branch

The high-resolution branch in our network is responsible for the spatial details. Using wide channels and shallow layers can provide sufficient spatial information. Inspired by the detail branch from BiSeNetV2, we design our own high-resolution branch. It uses three convolution layers consisting of 3×3 convolutions for channel expansion, and engages a max-pooling after each layer to quickly down-sample the input image to a scale of 1/8. Since the high-resolution branch only needs to focus on local details and computational overhead also needs to be controlled, neither dilated convolutions nor residual connections are involved in the whole process. At last, the output feature map not only preserves the edge details of the objects in the image but also reduces the resolution and expands the channel simultaneously. Even though the semantic extraction effect of this branch is not satisfactory, its main function is to transfer spatial information to the low-resolution branch in subsequent computation, so the feature extraction capability is not discussed on this branch.

### B. Low-resolution Branch

The low-resolution branch in our network is responsible for the semantic information. The basic construct of this branch is the

Residual U-blocks from U2-Net. RSU block is a U-Net[16] like symmetric encoder-decoder structure with height of *L*. The structure of RSU-*L*(*L*=7) ($C_{in}$, *M*, $C_{out}$) is shown in Fig. 3, where *L* is the number of layers in the encoder, $C_{in}$, $C_{out}$ denote input and output channels, and *M* denotes the number of channels in the internal layers of RSU. To any scale of RSU, the input feature maps and the output feature map has the same resolution. In our low-resolution branch, we linearly stacked six RSU modules of different sizes. Hence our low-resolution branch consists of six stages, each stage is filled by a well configured RSU block, the scales of *L* are (7,6,5,4,4,4). After each RSU block, we connect a max-pooling layer with a stride of 2, except the last stage. The last stage will output feature map with 1/32 resolution, we connect a context embedding[7] module after it to embed the global contextual information. Then the feature maps will be 2× up-sample and add up to the output of stage5. Finally, we will get the feature maps with 1/16 resolution of input image as our low-resolution branch's output. The design of our low-resolution branch inherits the advantages of U²-Net. After the refinement by us, this branch allows having deep architecture with rich multi-scale features and relatively low computation and memory costs.

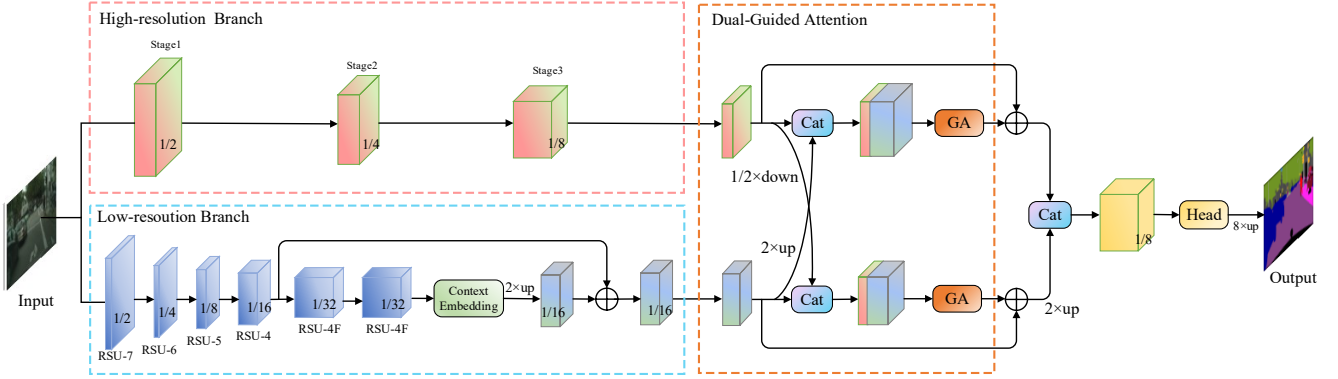

Fig. 2. **The overall architecture of our proposed BiDGANet.** There are mainly three parts: The High-resolution Branch in the pink dashed box, the Low-resolution Branch in the blue dashed box, and the Dual-Guided Attention module in the orange dashed box. The High-resolution Branch have three stages(the pink cubes) while the Low-resolution Branch has six RSU blocks(the blue cubes). Numbers in the cubes are the feature map size ratios to the resolution of the input. The specific scales of channel number are shown in Table 1. The last stage of the Low-resolution Branch is the output of the Context Embedding Block, we adopt a green cube to represent it and the bottom of blue cube gradients to green after this stage. In the Dual-Guided Attention part, "1/2×down" indicates the down-sampling operation, "2×up" represents the up-sampling operation, blue rounded rectangle "Cat" means concat operation, "GA" means Guided Attention. As shown in the figure, we design a parallel structure to realize our DGA module.

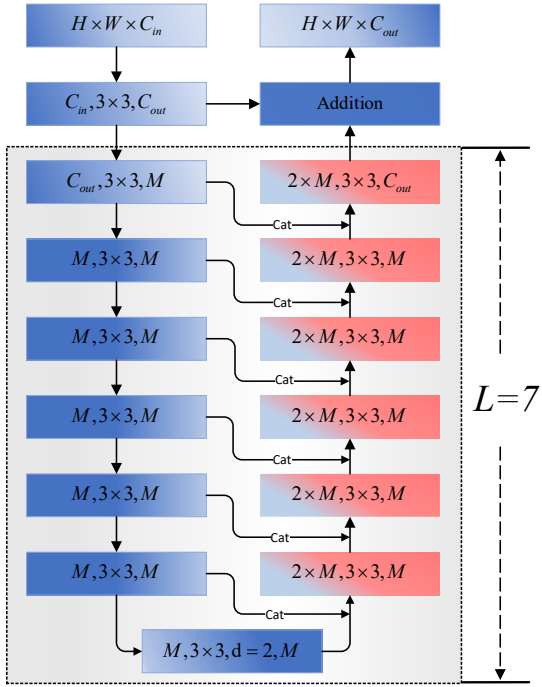

Fig. 3. **The overview of t Residual U-block while *L*=7.** The blue gradient rectangle represents a series of operations includes convolution, batch normalization, ReLU and max-pooling. Meanwhile the pink gradient rectangle respresents the similar operations, the differences are the number of input and output channels and the up-sampling operation replaces max-pooling. The other scales of RSU blocks mainly the difference in the height of *L*, and the number of input and output channels.

### C. Dual-Guided Attention

Inspired by External Attention (EA)[17], we realize that linear complexity attention mechanism can be introduced to real-time semantic segmentation and will not cause performance degradation. As shown in Fig. 4, we make some modifications on the original EA, and name it as Guided Attention(GA). Based on this, we propose the Dual-Guided Attention module, which is shown as the orange dashed box in the right part of Fig. 4. The module first uses a parallel structure to downsample the output feature map of 1/8 image resolution from the high-resolution branch to half of its own size by through a 3×3 convolutional layer and upsample the feature map of 1/16 image resolution from the low-resolution branch to twice times of itself by linear interpolation. Then both of the feature maps concatenate themselves to the other branch's output. After the concat operation, we get two feature maps with different resolutions but the same number of channels for Guided Attention computation separately. We will get into the technical details of GA later, but all we need to know here is that the output of a GA has the same number of channels as the output of a branch with the same resolution. This is mainly achieved through the Dropout layer in the final stage of GA. Therefore, we directly add the output of GA to the output of each branch at the corresponding resolution and implement a residual structure. Finally, we upsample the lower resolution feature map twice times and concatenate it with the higher resolution feature map, after which we send the final feature map into the segmentation head to get the segmentation prediction.

The details of Guided Attention are shown in Fig. 4. Given an input feature map $F \in \mathbb{R}^{N \times d}$, where the *N* is the number of pixels in images and *d* is the number of feature dimensions. According to the design of External Attention (EA), we also set up two fully connected layers as external learnable units separately. They are described as $M_k$, $M_v$, $\in \mathbb{R}^{S \times d}$, while $S$ is the number of nodes in the fully connected layer. In this article $S$ is set to 64 as a hyperparameter. $M_k$ and $M_v$ serve as the Key and Value in the attention calculation, so is not hard to know the input feature map $F$ would be Query. We briefly describe the calculation of GA as:

$$A = Norm(F_{in} \otimes M_k^T) \quad (1)$$

$$F_{mid} = A \otimes M_v^T \quad (2)$$

$$F_{out} = Drop(F_{mid}) \quad (3)$$

The $\otimes$ in the formula (1) and (2) means the inner product between input feature map $F_{in}$ and transpose of external learnable units $M$. To avoid the problem that attention map $A$ is sensitive to the scale of the input features and let our GA only be affected by the number of dimensions of the data from the feature map. We employ the double-normalization proposed in [18], which separately normalizes columns and rows. This double-normalization is the *Norm* operation in formula (1) and formulated as:

$$\tilde{A} = F_{in} \otimes M_k^T \quad (4)$$

$$\hat{a}_{i,j} = \frac{\exp(\tilde{a}_{i,j})}{\sum_k \exp(\tilde{a}_{k,j})} \quad (5)$$

$$a_{i,j} = \frac{\hat{a}_{i,j}}{\sum_k \hat{a}_{i,k}} \quad (6)$$

The matrix $\tilde{A}$ is the output of formula (4), it is the original output of first step calculation and the size is $(k, k)$. The next formula means we do a SoftMax calculation at each row of the matrix $\tilde{A}$, the output of each element is $\hat{a}_{i,j}$. After that, we do a L1-Normalization at each column of the matrix.

At the end of each GA, we use a Dropout layer. On the one hand, it can reduce the number of channels instead of 1×1 convolution, and on the other hand, it can prevent overfitting. With the Dropout, we can separately get two tensors that have the same size as the outputs of the two branches in the previous stage, and make the residual calculation directly.

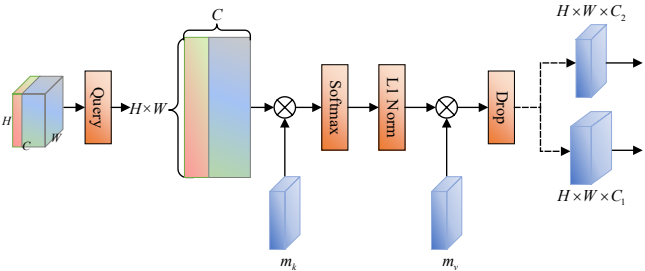

Fig. 4. **Illustration of Guided Attention**. This module refines the original external attention. The input is concatenation of two branches, we resize the tensor to a matrix, then comes the first product between the input matrix and the $m_k$ units. After that is the double-normalization. Next, the second product between the normalized matrix with the $m_v$ units. At last, we set the coefficient of the dropout layer to get the output we want.

## IV. EXPERIMENTS

We have conducted experiments on two standard benchmarks: Cityscapes[11] and Cam Vid[12]. We first introduce the datasets and the implementation details. Next, we conduct ablation experiments on Cityscapes validation set to demonstrate the proposed module's validity. Finally, we compare our network with other methods.

### A. Datasets and Evaluation Metrics

Cityscapes focuses on semantic understanding of urban street scenes from a driving perspective. The fine annotated images of this dataset are split into training, validation and test sets, with 2975, 500 and 525 images respectively. All the images have the same size which is 1024×2048 resolution. In our experiments, we follow the 19 semantic categories standard instead of the original 34 categories. We only use the fine annotated images to train and validate our network.

Cambridge-driving Labeled Video Database (CamVid) is also a street scene dataset for the real-time Semantic Segmentation, it contains 701 densely annotated frames and the resolution of each frame is 720 × 960. These frames are divided into 367 training images, 101 validation images, and 233 testing images. CamVid have 32 categories which has the subset of 11 classes are used for segmentation task.

For all the methods involved in the experiment, as well as the model proposed in this paper, we adopt mean of class-wise intersection over union (mIoU) and Frames Per Second (FPS) as the evaluation metrics. However, some of contrastive methods, whose FPS cannot be accurately measured by the scripts we write, we evaluate their performance by the average inference delay of a single image, the measuring unit would be millisecond(ms).

### B. Implementation Details

**Training settings**. Since our model does not have pre-training weights, we need to train our model on the train set of both Cityscapes and CamVid from scratch. The optimizer we choose is the SGD algorithm, the momentum is set to 0.9. And we also adopt the warm-up strategy and the "poly" learning rate scheduler, based on the number of iterations, we set the warm-up iterations to 1/100 of the number. On Cityscapes datasets, we randomly crop the input images to 512×512 resolution so that the train set can be extended and the batch size can be set to 32 on a single Nvidia Quadra A6000 GPU which has 48 gigabytes video memory. For learning rate, we use 0.01 as the initial setting and $e^{-4}$ as the weight decay in the optimizer. The number of epoch is set to 1000 in order for the model to learn sufficiently. On CamVid datasets, the cropped size changes to 672×672, and the batch size is set to 16. Due to the lack of sufficient training images, we set 0.005 as the initial learning rate and $5e^{-5}$ as the weight decay. Also, the epoch is set to 400 which is less than that on Cityscapes in order to avoid overfitting on the training set.

**Inference settings**. For a fair comparison, we performed all inference experiments under CUDA 12.0, CUDNN 8.7, and without TensorRT for acceleration on a single Nvidia RTX 3070 GPU. For all experiments, the batch size of inference is set to 1. Our method uses full resolution of both datasets without image resizing. As mentioned in the previous section, some methods for comparison acquire the scaled images, generate the predicted images. These operations make our scripts that calculate FPS cannot get their FPS data accurately. For these methods we show the average inference delay of one single image which include those resize steps, and calculate approximate FPS without the resize steps. At the same time, we also show our method's average inference delay of singe image.

## C. Ablation experiments on Cityscapes

In the following experiments, we adopt our large version network to validate the effectiveness of each component in our method, because it has the the largest number of parameters which would show the improvements more obviously. The Table I shows the three versions of our network, the number of channels in High-resolution branch is same and the main difference is the number of channels in the Low-resolution branch.

TABLE I.
THE DIFFERENT VERSIONS OF OUR NETWORK. ALL VERSIONS HAVE SAME SCALE OF HIGH-RESOLUTION BRANCH. "RSU" DENOTES RESIDUAL U-BLOCK, "I", "M" AND "O" INDICATE THE NUMBER OF INPUT CHANNELS ($C_{IN}$), MIDDLE CHANNELS AND OUTPUT CHANNELS ($C_{OUT}$) OF EACH BLOCK.

| Version | High-resolution Branch | | | Low-resolution Branch | | | | | |
|---|---|---|---|---|---|---|---|---|---|
| | Stage1 | Stage2 | Stage3 | RSU-7 | RSU-6 | RSU-5 | RSU-4 | RSU-4 | RSU-4 |
| Light | I: 3 | I: 64 | I: 64 | I: 3 M:16 O:32 | I:32 M:16 O:64 | I:64 M:16 O:64 | I:64 M:16 O:64 | I:64 M:16 O:64 | I:64 M:32 O:64 |
| Base | O: 64 | O: 64 | O: 128 | I: 3 M:16 O:32 | I:32 M:16 O:64 | I:64 M:32 O:128 | I:128 M:64 O:256 | I:256 M:128 O:256 | I:256 M:128 O:256 |
| Large | | | | I: 3 M:32 O:64 | I:64 M:32 O:128 | I:128 M:64 O:256 | I:256 M:128 O:512 | I:512 M:256 O:512 | I:512 M:256 O:512 |

*a) Two branches:* We explore the work of individual branch first. As shown in the first three rows of Table II, the high-resolution branch and low-resolution branch alone achieve 43.8% and 66.2% mIoU. However, with just one concat operation, the final outputs of the two branches can achieve 70.6% mIoU.

*b) External Attention (EA):* Using both branches, we achieve considerable improvement by directly following an EA module proposed in paper [17], which shows its effectiveness.

*c) Dual-Guided Attention (DGA):* Based on our DGA module, we design the network shown in Fig 1. With our DGA module, low-level spatial details and high-level semantics interact each other and selectively fuse in this module. Compared to just using a single EA, it brings a nearly 4% improvement. At last, we use online hard example mining[19](OHEM) to replace the basic cross-entropy loss and gain a little improvement.

TABLE II.
ABLATION STUDY OF OUR LARGE VERSION NETWORK ON CITYSCAPES.

| High-resolution | Low-resolution | Components | | | OHEM | mIoU(%) |
|---|---|---|---|---|---|---|
| | | Concate | EA | DGA | | |
| ✓ | | | | | | 43.8 |
| | ✓ | | | | | 66.2 |
| ✓ | ✓ | ✓ | | | | 70.6 |
| ✓ | ✓ | ✓ | ✓ | | | 73.1 |
| ✓ | ✓ | ✓ | | ✓ | | 76.8 |
| ✓ | ✓ | ✓ | | ✓ | ✓ | 77.9 |

## D. Experiments on Cityscapes

We compare our network to previous real-time methods in the experimental configuration described above. Table III shows each method's model information, includes the input resolution size, the mIoU, the GPU device, and the frames. Some of these methods we follow its original training setting and evaluate their performances on our device, we mark them with symbol *. All other data are extracted from the original papers of these methods. The experimental evaluation indicates that our method shows a competitive result. The light version has good speed performance, and the basic version has a good trade-off between speed and accuracy, while the large version has the accuracy to compete with the state-of-the-art methods.

## E. Experiments on CamVid

We also evaluate our method on the CamVid dataset. The original resolution of 960×720 is used for all the methods evaluated on CamVid, so the resolutions are not listed in Table IV. As same as Table III, the symbol * means we repeat the result on our device and environment setting. Results on the CamVid dataset show our method also achieves excellent performance, which means our method can adapt to different picture qualities and has good generalization ability.

TABLE III.
ACCURACY AND SPEED COMPARISON ON CITYSCAPES VAL SET.

| Method | Resolution | GPU | mIoU (%) | Frames (FPS) | Time (ms) |
|---|---|---|---|---|---|
| ICNet*[5] | 2048×1024 | RTX 3070 | 68.9 | 58 | -- |
| STDC$_2$-Seg50*[23] | 1024×512 | RTX 3070 | 73.5 | -- | 11 |
| STDC$_2$-Seg75*[23] | 1536×768 | RTX 3070 | 76.3 | -- | 16 |
| BiSeNet V2*[7] | 2048×1024 | RTX 3070 | 73.1 | 143 | -- |
| BiSeNet V2-L*[7] | 2048×1024 | RTX 3070 | 75.6 | 41 | -- |
| DDRNet-23-silm*[15] | 2048×1024 | RTX 3070 | 77.2 | -- | 12 |
| DDRNet-23*[15] | 2048×1024 | RTX 3070 | 78.8 | -- | 22 |
| DFANet A[10] | 1024×1024 | Titan X | 71.3 | 100 | -- |
| DFANet B[10] | 1024×1024 | Titan X | 67.1 | 120 | -- |
| PP-LiteSeg-B$_1$[24] | 1024×512 | GTX 1080ti | 73.9 | 196 | -- |
| PP-LiteSeg-B$_2$[24] | 1536×768 | GTX 1080ti | 77.5 | 102 | -- |
| PIDNet-L(sota) [25] | 2048×1024 | RTX 3090 | 80.6 | 31 | -- |
| RTFormer (sota) [26] | 2048×1024 | RTX 2080ti | 79.3 | 39 | -- |
| BiDGANet-Light(our) | 2048×1024 | RTX 3070 | 71.1 | 165 | 6 |
| BiDGANet-Base(our) | 2048×1024 | RTX 3070 | 75.2 | 73 | 14 |
| BiDGANet-Large(our) | 2048×1024 | RTX 3070 | 77.9 | 43 | 21 |

TABLE IV.
ACCURACY AND SPEED COMPARISON ON CAMVID TEST SET.

| Method | GPU | mIoU (%) | Frames (FPS) |
|---|---|---|---|
| ENet[27] | -- | 51.3 | -- |
| ICNet[5] | Titan X M | 67.1 | 35 |
| DFANet A[10] | Titan X | 64.7 | 120 |
| BiSeNet V1[6] | Titan X | 65.6 | 175 |
| BiSeNet V1-L[6] | Titan X | 68.7 | 117 |
| BiSeNet V2*[7] | RTX 3070 | 72.4 | 33 |
| BiSeNet V2-L*[7] | RTX 3070 | 73.3 | 110 |
| STDC$_1$-Seg[23] | GTX 1080ti | 73 | 197 |
| STDC$_2$-Seg[23] | GTX 1080ti | 73.9 | 152 |
| S$^2$-FPN18[28] | GTX 1080ti | 69.5 | 107 |
| S$^2$-FPN34[28] | GTX 1080ti | 71.1 | 124 |
| PP-LiteSeg-T[24] | RTX 2080ti | 73.3 | 222 |
| PP-LiteSeg-B[24] | RTX 2080ti | 75.0 | 154 |
| BiDGANet-Base(our) | RTX 3070 | 75.9 | 103 |

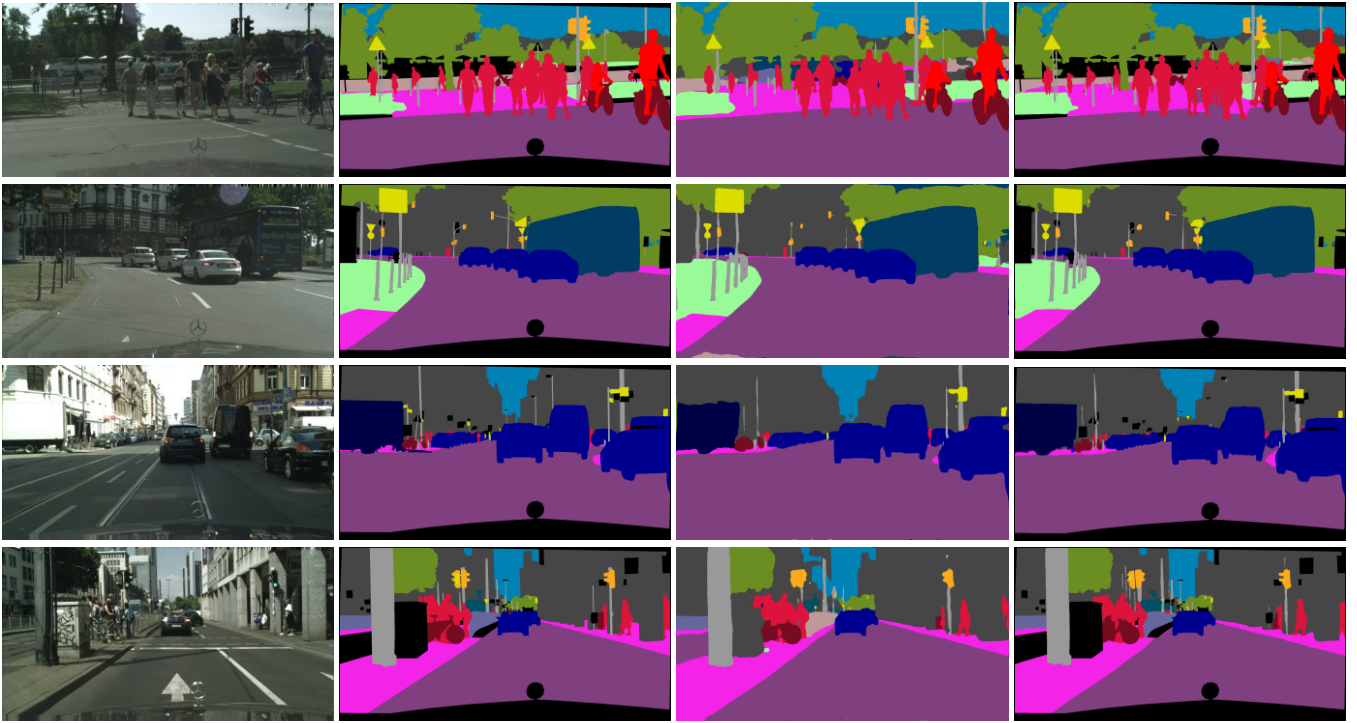

Fig. 5. Visualization of segmentation results on the Cityscapes val set. The four columns left-to-right refer to the input image, the ground truth, the output of BiDGANet-Light, the output of BiDGANet-Large. We mask the classes that do not participate in the segmentation on the Large version as a distinction. All images are in the original 2048×1024 resolution.

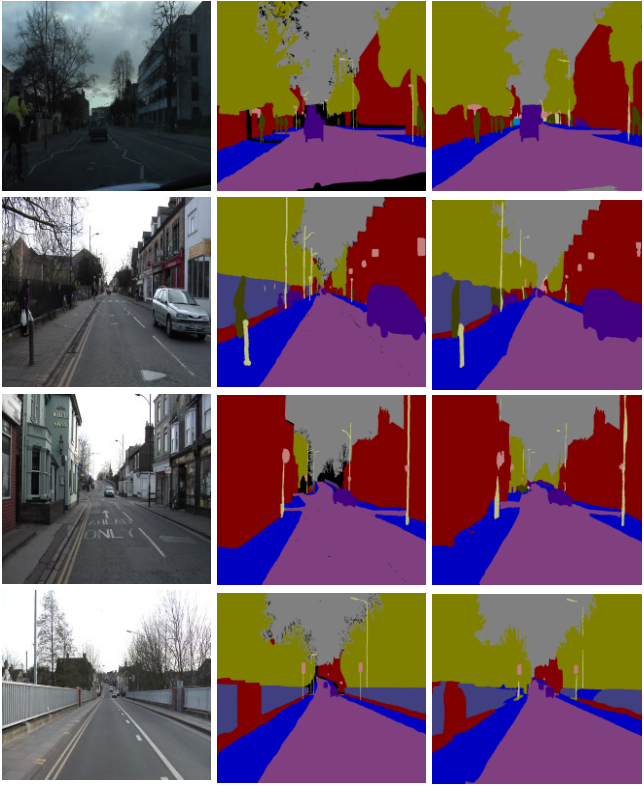

Fig. 6. Visualization of segmentation results on the CamVid test set. The three columns left-to-right refer to the input image, the ground truth, the output of BiDGANet-Base.

## V. CONCLUSIONS

In this paper, we present a new two-branch architecture network based on Residual U-blocks and Dual-Guided Attention. To our best knowledge, we are the first to introduce the RSU block into real-time semantic segmentation, this brilliant design has already shone in the salient object detection task and with this block, we design a powerful multi-scale feature extraction network as our low-resolution branch to replace the hand-craft one of our baseline. Ultimately, with the new Dual-Guided Attention module we proposed, our method achieves competitive results on two popular benchmarks. In conclusion, as the initial research observations and positions, we basically meet the anticipative desire. Meanwhile, due to the simplicity and flexibility of our method, it can be further improved according to different conditions. Further studies will focus on its more structural possibilities with Neural Architecture Search(NAS) technology, and make efforts to develop it into a baseline in the future.